\newcommand{\paragraphb}[1]{\vspace{0.75ex}\noindent{\bf #1}\hspace*{.3em}}

\documentclass[letterpaper,twocolumn,10pt]{article}
\usepackage{usenix,epsfig,endnotes}
\usepackage{amsmath}
\usepackage{amssymb}
\usepackage{algorithm,algorithmic}
\usepackage{subcaption}

\begin{document}


\date{}

\title{\Large \bf FaceLeaks: Inference Attacks against Transfer Learning Models via Black-box Queries}

\author{
{\rm Seng Pei Liew} \\
LINE Corporation
\and
{\rm Tsubasa Takahashi} \\
LINE Corporation
}

\maketitle

\thispagestyle{empty}

\subsection*{Abstract}
Transfer learning is a useful machine learning framework that allows one to build task-specific models (student models) without significantly incurring training costs using a single powerful model (teacher model) pre-trained with a large amount of data.
The teacher model may contain private data, or interact with private inputs. We investigate if one can leak or infer such private information without interacting with the teacher model directly.
We describe such inference attacks in the context of face recognition, an application of transfer learning that is highly sensitive to personal privacy.

Under black-box and realistic settings, we show that existing inference techniques are ineffective, as interacting with individual training instances through the student models does not reveal information about the teacher. We then propose novel strategies to infer from aggregate-level information. Consequently, membership inference attacks on the teacher model are shown to be possible, even when the adversary has access only to the student models.

We further demonstrate that sensitive attributes can be inferred, even in the case where the adversary has limited auxiliary information. Finally, defensive strategies are discussed and evaluated.
Our extensive study indicates that information leakage is a real privacy threat to the transfer learning framework widely used in real-life situations.

\section{Introduction}
Recent years have seen an exponential growth in the field of machine learning (ML), particularly learning based on deep neural networks (DNNs). These technologies are now widely applied across industry spanning from image recognition \cite{DBLP:conf/nips/KrizhevskySH12}, natural language processing \cite{devlin2019bert}, speech recognition \cite{hannun2014deep}, and even to high-stake applications such as medical diagnosis \cite{choi2016doctor}. 

The keys to these successes are not only limited to advances in algorithms and architectures; the availability of large datasets contributes to building better and more accurate models as well. As ML models are increasingly used as part of decision making in various critical applications, the research community is beginning to study the interplay between ML algorithm and data to ensure fairness, privacy, and transparency in critical decision making processes.

To give a few examples: commercial face recognition systems are found to be discriminative with respect to classes such as race and gender \cite{DBLP:conf/fat/BuolamwiniG18}.
It is demonstrated that privacy can be leaked from ML models: one can deduce whether an instance belongs to the membership of the training data based on ML prediction outputs \cite{DBLP:conf/sp/ShokriSSS17}. 
Such studies motivate the need to exercise caution when deploying ML applications in practice.

\paragraphb{Transfer learning.} In this work, we concentrate on privacy issues related to transfer learning.  
Transfer learning is a paradigm that seeks to transfer knowledge (gained from existing domains) to accomplish tasks in a new domain. 
As training large and accurate models (involving millions of parameters for the case of deep learning) requires a large amount of domain-specific data, it is costly resource-wise to train independently different models for specific (downstream) tasks. 
Transfer learning seeks to resolve this using a single, transferable, pre-trained and task-agnostic model.

From the perspective of ML practitioners, transfer learning also allows them to focus on collecting data solely for training the teacher model. 
Then, task-specific student models trained using user-provided data may be deployed and exposed to users as a service via APIs.

While it is known that exposing ML models directly to users can cause privacy leakage (e.g., \cite{DBLP:conf/sp/ShokriSSS17}), it is also vital to carefully inspect whether the underlying data used to train the teacher model can be leaked within the transfer learning framework, even though the teacher model is not exposed to users directly.
We will investigate this issue in depth in this work.

\paragraphb{Face recognition.} We emphasize the privacy leakage issues by focusing primarily on face recognition within the transfer learning framework in this work. The reasons are three-fold.
First, modern face recognition models leverage transfer learning to various degrees \cite{DBLP:conf/bmvc/ParkhiVZ15, DBLP:conf/nips/SunCWT14,DBLP:conf/cvpr/TaigmanYRW14,DBLP:conf/cvpr/WangWZJGZL018} and therefore serve as a prime use case of transfer learning.
Second, some of the state-of-the-art face recognition results are achieved relying on the use of private data (e.g., \cite{DBLP:conf/cvpr/WangWZJGZL018}), necessitating considerations of privacy. 

Third, while face recognition models are now widely deployed in our society, ranging from immigration inspection 
to smartphone authentication,
privacy issues related to face recognition remains a hotly debated topic.
Countries such as the UK, China and Singapore are scanning the faces of millions of citizens without consent \cite{Aravindan.2018}.
A private company called Clearview.ai is reported to have collected billions of photographs online to train large face recognition models that are capable of identifying millions of citizen without explicit consent \cite{Hill.2020}.
It is therefore timely to study from various perspectives the privacy issues of face recognition.

\paragraphb{MLaaS and privacy.} Service providers commercializing this technology will be concerned about privacy issues arisen by it; the privacy of the teacher model can be important as the data collection process may be expensive and the collected data may be private. 

Particularly, there are currently multiple companies providing ML as a service (``MLaaS"), including internet giants such as Google and Amazon. 
These companies usually impose privacy policies when collecting personal information and data to protect individual's privacy. Leakage of information about training data may then undermine the reputation of the company.
Moreover, the rise of privacy laws means that legal issues may arise when MLaaS providing companies train models using data collected by, e.g., Clearview.ai, which contain personal data collected without consent; companies violating the laws may face hefty fines.

Furthermore, the adversary who has learned the private information may misuse it in various ways.
Stalkers may use such information to figure out the leaked identity via social media platforms.
Leaked face images may raise identity theft concerns as well \cite{Cross.2019}.

The present work is thus of interest to MLaaS providing companies, as it serves as a way to assess the risk of disclosing the details of training data. Additionally, perhaps more importantly, our work has impact on common citizens concerned with their privacy: Our methodology can help figure out if one's sensitive information is being exploited without consent. 
We believe that our tools are important for common citizens to protect themselves in today's world where privacy violations are increasingly prevalent.

\paragraphb{Our contributions.}
We launch a systematic study on potential privacy leakage issues in face recognition systems which utilize transfer learning.
We consider realistic scenarios where the adversary has black-box queries to the face recognition system in various real-life scenarios. 
We also introduce novel techniques to attack models, extending existing attacks and threat models. 
Our results are based on widely-used face recognition methodologies
\cite{DBLP:conf/cvpr/LiuWYLRS17, DBLP:conf/cvpr/WanZLC18, DBLP:journals/spl/WangCLL18, DBLP:conf/cvpr/WangWZJGZL018,DBLP:conf/cvpr/DengGXZ19}. 
The highlights of our contributions are as follows:
\begin{itemize}
    \item It is possible to perform membership inference attacks on teacher models \textit{solely} with black-box accesses to the student models. The success metric, measured with the area under the curve (AUC), is as high as 0.71.
    \item Attribute inference attacks are possible through the APIs. Non-trivial inference (AUC larger than 0.5) is possible under constrained conditions (e.g., under limited auxiliary information).
    \item Mitigation is partially viable, albeit it comes at a price of utility, e.g., injecting noises introduces uncertainty, revealing less information hinders one from knowing how confident a model is on a sample. 
\end{itemize}

\paragraphb{Paper organization.}
The rest of the paper is organized as follows. 
Section \ref{sc:bkg} provides necessary background for the rest of the work.
The problem statement, threat models considered, and setups are described in Section \ref{sc:prb}.
Membership inference attacks are dealt with in Sections \ref{sec:stra} and \ref{sec:eva}.
Section \ref{sec:attr} covers attribute inference attack.
Section \ref{sec:defense} describes possible defenses against such attacks.
We provide works related to this paper in Section \ref{sec:related} before concluding in Section \ref{sec:conclusion}.

\section{Background}
\label{sc:bkg}
In this Section, we lay out background information needed to further our discussion of inference attacks against transfer learning.
 
\subsection{Transfer Learning}
We here discuss minimal concepts required for the incoming sections of the paper regarding to transfer learning. 
A complete survey of transfer learning can be found in \cite{DBLP:journals/tkde/PanY10}.

At the high level, transfer learning frameworks aim to pass the knowledge of a pre-trained teacher model to student models.
For the case of convolutional neural networks (CNNs), which is widely use in the image recognition domain, it is understood that lower convolutional layers are able to capture low-level image features, such as edges, while higher convolutional layers can capture more complex features, such as face attributes \cite{DBLP:conf/nips/KrizhevskySH12}.
The final layers of a CNN are understood to be able to capture enough information to solve the task assigned to the teacher.

In order to solve a new task which bears similarity to the task of the teacher, it is then reasonable to extract features from the final layers of the teacher model, which contains knowledge such as face attributes and edge combinations.
The model under such setups is referred to as \textit{feature extractor}, of which we denote $\psi$.
The knowledge or feature is typically represented mathematically as a vector of real numbers.

\begin{figure}[t]
    \centering
    \includegraphics[width=7cm]{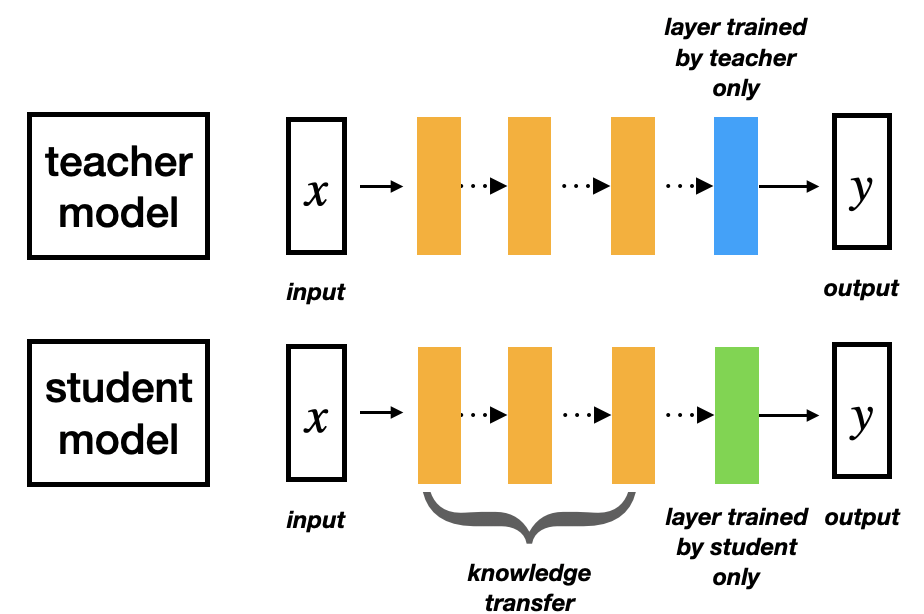}
    \caption{Transfer learning. A teacher model is first pre-trained with its tasks. Then, the intermediate layers are used by the student model to train its model-specific tasks.}
    \label{fig:trans}
\end{figure}

The knowledge of the teacher model is passed to a student model, $S_{\theta}$ (where $\theta$ represents the model weights specific to the student model), via $\psi$. $S_{\theta}$ can be as simple as a step function learning the decision threshold, or can be a slightly complicated neural network model with a few additional layers. Typically, the model weights of $\psi$ are ``frozen" or tuned at a lower learning rate while training $S_{\theta}$ such that the pre-existing teacher knowledge are kept and can be passed to the student. 
Figure \ref{fig:trans} illustrates the transfer learning approach we are interested in.

\subsection{Face recognition}
There are mainly two categories of face recognition tasks. First, there is the \textit{closed-set face classification}, which can be formulated as a supervised multiple-class classification problem, where the the objective is to determine which identity a face image belongs to among a fixed number of classes of identity. 
Classifying new identities or classes not seen by the teacher model can be efficiently performed by the student model via transfer learning. 
Such a scenario is called \textit{fine-tuned face recognition} in this work.

The second category of face recognition is known as face verification or \textit{open-set face recognition}. 
In this scenario, a candidate face image is to be compared with a pre-determined set of face images to check if the candidate matches one of the identities to be compared with.

\paragraphb{Face recognition by feature extractors.}
While the close-set face classification is straightforward a multiple-class classification task, the open-set face recognition task is different. Modern face recognition systems usually involve transfer learning, i.e., adopting a pre-trained DNN-based feature extractor to map face images to a low-dimensional feature vector \cite{DBLP:conf/bmvc/ParkhiVZ15, DBLP:conf/nips/SunCWT14,DBLP:conf/cvpr/TaigmanYRW14,DBLP:conf/cvpr/WangWZJGZL018}. Then, the similarity between the two face images can be quantified to decide whether they belong to the same identity or not. 

There are mainly two methods of pre-training the teacher DNN-based feature extractors.
The feature extractor can be trained directly via metric learning \cite{DBLP:conf/cvpr/SchroffKP15}. This is achieved through the triplet loss, where a triplet is formed using two matching face images and a non-matching face image. The training objective is to separate the positive pair from the negative one in the feature space.
This approach has several shortcomings, including scalability, as the number of triplet combination increases exponentially with the number of training data points. 
Moreover, the training involves semi-hard mining, which is known to be difficult to implement.

Another approach of training the teacher feature extractor is to initially train a standard multi-class DNN classifier to classify the face images in the training data. 
Then, the output of the penultimate layer of the DNN is treated as the feature to be used for transfer learning.
We adopt this approach in this work due to the difficulties of metric learning mentioned above. It should be noted that more recent state-of-the-art works adopt this approach to obtain better performance as well \cite{DBLP:conf/cvpr/LiuWYLRS17, DBLP:conf/cvpr/WanZLC18, DBLP:journals/spl/WangCLL18, DBLP:conf/cvpr/WangWZJGZL018,DBLP:conf/cvpr/DengGXZ19}. 

\paragraphb{Face recognition in practice.}
Finally, we describe several face recognition services available in the commercial market.
The Microsoft Azure API provides a service to identify a detected face against a database of people. 
Within the API, there is a  ``training endpoint” that must be called before face recognition is initiated. Transfer learning is likely to be used by Microsoft to train a face recognition model on user-submitted images.

Both the Microsoft Azure API and the Amazon Rekognition service are able to perform face verification: they provide services to check the likelihood of two user-submitted faces images belonging to the same person and output a score.

Other companies providing similar services include SkyBiometry, Kairos, and Lambda Labs.

\subsection{Legal issues}
From a legal point of view, our domain of interest, face recognition, is a form of biometrics where laws and regulations protecting this kind of privacy are well established.
The EU General Data Protection Regulation (GDPR) treats biometric data as a subject of regulation and requires details of a breach to be reported within 72 hours in case a breach occurs.
The Japanese Act on the Protection of Personal Information also states clearly that biometric data such as fingerprints and facial attributes are personal and the subject must be notified when such data are collected and utilized. 
Though, the regulation on ML applications handling biometric data remains subtle.

\section{Problem statement}
\label{sc:prb}
We give an overview of the problem in consideration in this Section.

MLaaS providers pre-train a private model and provide other parties the output features for downstream tasks.
It is tempting to think that it is safe and private as the model is not shared; merely a vector of floating-point numbers or confidence scores are exposed to potential adversaries.
However, sensitive information might be embedded in these ``numbers" and could be extracted.

Our research questions are, what kinds of sensitive information can an adversary learn by interacting indirectly with a private model?
And what can an adversary infer about the input to the model, which can be private, by interacting with the corresponding API? 

\subsection{Threat models}

We assume that the adversary does not have the knowledge of the details of the teacher model. Information about the architecture used and model internals are not available, i.e., we consider black-box inference attacks throughout this work.
The same applies to the student models.

\begin{figure*}
    \centering
    \includegraphics[width=0.9\textwidth]{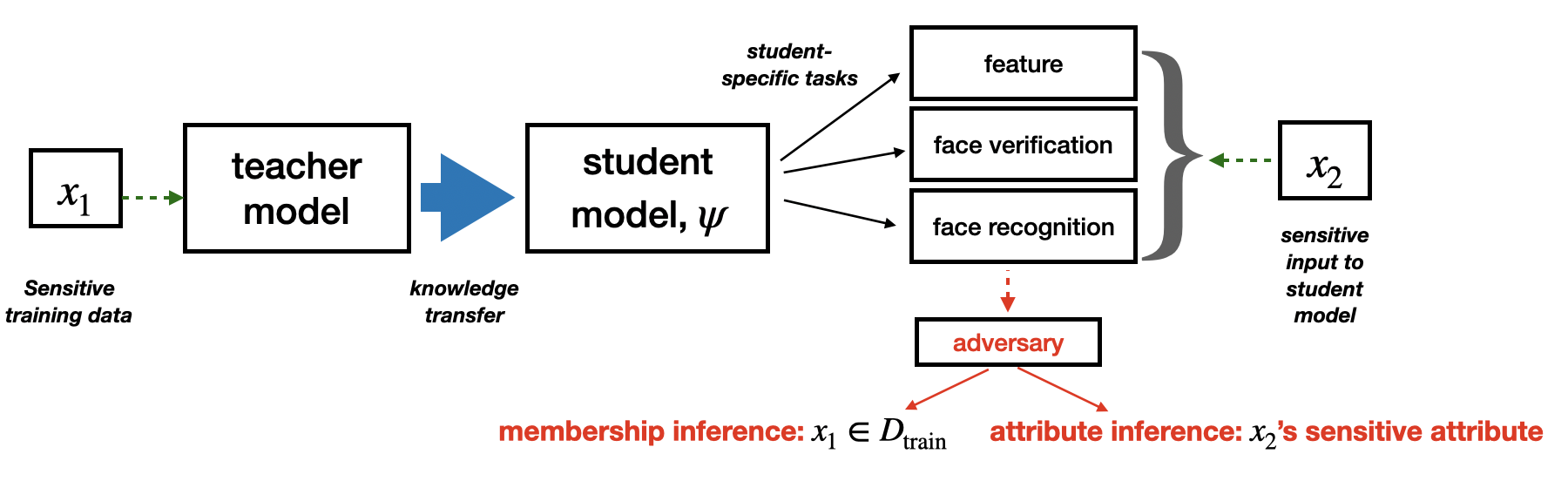}
    \caption{An overview of the transfer learning setting and attacks in consideration. The adversary is assumed to be only interacting with the student models, and launch inference attacks to (1) infer the membership, i.e., whether the input $x_1$ has been used in training the teacher model, (2) infer sensitive attributes of $x_2$ without access to it.}
    \label{fig:overview}
\end{figure*}

The overview of our setting and attacks is illustrated in Figure \ref{fig:overview}.
The adversary has query access only to the student models' output (APIs), and perform inference attacks based on the responses of the APIs.

The three types of APIs that the adversary interacts with are as follows (with a slight abuse of notation, we reuse $\psi$ to denote all types of APIs of interest).

\paragraphb{DNN features.} The adversary is assumed to be able to access the feature vectors of the teacher model.
This is the case where the student model can be downloaded by the user to run it locally, albeit the user needs to make queries to obtain the features as inputs to the student model.

\paragraphb{Face verification.} Two face images are queried at a time. In return, the adversary gets a similarity score from the API.

\paragraphb{Fine-tuned face recognition.}
A face image is queried, and the API responds by outputting a vector of prediction probabilities  (with length equal to the number of classes to be classified) that the queried image belongs to a certain class.
We also call this the \textit{confidence scores}.

We formulate two classes of attack.

\paragraphb{Membership inference.} 
The teacher model contains training data that are sensitive.
The purpose of the adversary is to make inferences from $\psi$ about the dataset $D_{\rm target}$, which contains the training data, $D_{\rm member}$ and unrelated data, $D_{\rm non-member}$.
The adversary queries $D_{\rm target}$ to obtain API responses.

Auxiliary dataset may or may not be available. In the case where auxiliary dataset is not available, the adversary sorts the instances of $D_{\rm target}$ based on the score obtained from inference to figure out instances belonging to $D_{\rm member}$ at which she is most confident.

We note that membership inference attacks on the \textit{student model's training data} using the student model's output can be performed simply using techniques proposed in \cite{DBLP:conf/sp/ShokriSSS17} and will not be discussed further in this work. We are interested in membership inference attacks on the \textit{teacher model's training data} using the student model's output, which is a more subtle problem not studied before. 

\paragraphb{Attribute inference.}
The threat model is that an adversary wishes to learn the sensitive attributes, $s$ of secret inputs, $\textbf{x}$ into $\psi$ \cite{DBLP:conf/iclr/SongS20}.
We are interested in investigating how different interactions with APIs potentially leak sensitive information.

We assume that auxiliary dataset, $ (\textbf{x},s) \in D_{\rm aux}$ is available.
Since this reduces to the plain supervised learning given sufficient amount of data or  information from $\psi$, the more interesting case we would like to study is where labeled data is scarce, or the information obtainable from $\psi$ is limited.

Neither malicious service provider nor hostile client will be considered here.
These scenarios are important within the collaborative learning framework, but is out of the scope of this paper. 

\paragraphb{Metrics.} Next, we describe metrics used to measure the effectiveness of inference attacks.
In this paper, true positive (TP) denotes the number of correctly predicted data points belonging to the training dataset. False positive (FP) is the number of data points erroneously predicted as belonging to the training dataset, and false negative (FN) is the number of actual data points belonging to the training dataset wrongly predicted by the model. The metrics precision, recall and F1-score are used to gauge the performance.
They are defined as:
\begin{eqnarray}
{\rm Precision} &=& {\rm \frac{TP}{TP+FP}}, \nonumber \\
{\rm Recall} &=& {\rm \frac{TP}{TP+FN}},\nonumber \\
{\rm F1} &=& {\rm \frac{2\, Precision\cdot Recall}{Precision+Recall}}. \nonumber
\end{eqnarray}
The true positive rate (TPR) and false positive rate (FPR), defined as:
\begin{equation}
 {\rm TPR }= {\rm \frac{TP}{TP+FN}}, \,\, {\rm FPR }= {\rm \frac{FP}{FP+TN}}.\nonumber 
\end{equation}
are utilized to present our results with the receiver operating characteristic curve, or ROC curve. The ROC curve is a plot of TPR versus FPR with varying discrimination thresholds within the binary classification setting.
The area under the curve, AUC, is also important in this paper.
AUC represents the probability of a classifier to rank a randomly chosen positive instance higher than a randomly chosen negative instance.

\subsection{Teacher model}
Our implementation of DNNs is based on the PyTorch library.

The teacher model is trained via a pipeline consisting of a face detector followed by a face recognition module. 
The Multitask Convolutional Neural Network (MTCNN) is utilized to detect and crop face images \cite{DBLP:journals/spl/ZhangZLQ16}. 
A DNN trained with softmax loss and the Inception-Resnet-v1 architecture
serves as the face recognition module \cite{DBLP:conf/cvpr/HeZRS16,DBLP:conf/fgr/CaoSXPZ18}. 

The input to the DNN is an image rescaled to $160 \times 160$ pixels. The penultimate layer of the DNN extracts a
feature vector of 512 dimensions from face images. 
This output feature is the knowledge passed to the student models implemented in this work.

The teacher model is trained with the training subset of the VGGFace2 dataset, $\mathcal{D}_{{\rm VGG}_{\rm train}}$ to classify 8631 classes of face of $\mathcal{D}_{{\rm VGG}_{\rm train}}$ \cite{DBLP:conf/fgr/CaoSXPZ18}. 

\subsection{Student models}

\paragraphb{Face verification.}As one of the simplest form of student models, the face verification system simply outputs the Euclidean distance between the two feature vectors as the similarity score (Another common choice is the cosine distance).

A threshold on the score is pre-determined to verify whether two faces belong to the same identity or not.
The Labeled Faces in the Wild (LFW) \cite{LFWTech} dataset is used to evaluate the performance and determine the threshold.
We use the recommended pairs \cite{LFWTech} of individuals for both testing and thresholding.
A 10-fold cross validation on the LFW dataset is performed to determine the optimal decision threshold. 
We achieve an accuracy of 0.964 and the corresponding decision threshold is $\tau_{\text{LFW}}=1.232)$.

\paragraphb{Fine-tuned face recognition.}Student model is trained (fine-tuned) to perform face recognition on 20 identities/classes not pre-trained by the teacher model.
All weights up until the penultimate layers of the teacher model are kept constant during the fine-tuning phase.
We choose 20 identities/classes from the test subset of VGGFace2 with equal ratio of gender. 
10 images are selected from each identity/class.
A validation score of 99\% is achieved with 20\% of the selected images after training for 10 epochs.

A technical note on the batch normalization layer is in line. The mean and variance statistics are usually tracked and updated during training, and these running statistics are reused during inference. We strictly stop the tracking during the fine-tuning phase as well, although the results of inference attacks do not change much either way.

\section{Membership inference: strategies}
\label{sec:stra}
The general approaches of performing membership inference is given in this Section, before describing the concrete study of each API in the next Section.
In the following, we compare first-cut approaches, which are based on existing studies, with our proposed approaches.
We design the both methodologies which are tailored to the membership inference towards feature extractors.

\subsection{First-cut approaches}
The main idea of the first-cut approaches is that similar models trained on similar data should behave similarly.
Given auxiliary dataset $D_{\rm aux}$, one can train a classifier to determine the membership of the target dataset $\textbf{x} \in D_{\rm target}$, expecting that $D_{\rm target}$ behave similarly to  $D_{\rm aux}$.

The above statement is true if the labels of $D_{\rm aux}$ are obtained from the student model, and the adversary wishes to infer whether $D_{\rm target}$ belong to the \textit{same student model} as well.
However, our interest is inferring whether $D_{\rm target}$ belong to the \textit{teacher model} with $D_{\rm aux}$ labeled by the student model. 
$D_{\rm target}$ and  $D_{\rm aux}$ do not behave similarly as a result.
Empirical evaluation in the next Section shows that this first-cut approach does not produce meaningful inference results.

We also note that the adversary considered here is more powerful than those presented in the literature of membership inference attacks against ML.
In \cite{DBLP:conf/sp/ShokriSSS17}, shadow models are trained to infer the membership without direct access to the target model's training data, i.e., the attack model's training data and the target model's training data are disjoint.
Here, we will consider an adversary that is more powerful in the sense that she is able to train attack models directly with the target model's training data.
Even with this advantage, the adversary is unable to produce good inference results, as will be shown in the next Section.

\subsection{Class-based inference}
We propose novel strategies to improve the above solution and to better infer from the API responses. 
This is one of the main contributions of this work.

We propose to exploit the \textit{class} information of the data points as well, instead of only the membership information. 
Instead of considering the individual property of each instance, we hypothesize that the aggregate-level inference is much more informative.
This also makes more sense when considering privacy issues in the face recognition domain, as we are more interested in knowing if a person (class) has participated in the training, instead of knowing if a specific photo of a person is used in the training.

Our observation is that, feature vectors from the same class/identity that are in the training dataset are expected to be more ``concentrated" in the feature space.
This is because they are explicitly enforced by the training loss to be within the same boundaries.
Non-member data are not enforced to do so, and are expected to be less ``concentrated" in the feature space.
Deriving an aggregate-level measure for each class to capture this information is expected to help perform inference attacks.

In order to exploit the class information, the adversary must first deduce the class or identity of the face images.
This is in principle achievable by the adversary by training an auxiliary model, $\psi_{\rm aux}$, (e.g., using public data) to perform face recognition/verification.
Then, the aggregate-level similarity measure is derived for each identity/class $y_1$ and utilized to perform inference attacks.

Another advantage of the class-based inference compared to first-cut approaches is that one can perform meaningful inference attacks even \textit{without auxiliary knowledge}.
Fitting the samples in $D_{\rm target}$ with aggregate-level similarity measure (using, e.g., Gaussian functions) can provide useful information to the adversary.

We denote our approach \textit{class-based inference}, and as will be shown below, this approach is far more effective than the first-cut approaches.
The general class-based inference approach is described in Algorithm \ref{alg:class_based}.

\begin{algorithm}[t]
\caption{Class-based inference}
\begin{algorithmic}[1]
\label{alg:class_based}
\STATE \textbf{Input:} target dataset $\textbf{x} \in D_{\rm target}$ , target model $\psi$,  auxiliary model $\psi_{\rm aux}$, optional auxiliary dataset $D_{\rm aux}$
\STATE Label the class $\{y_1 = \psi_{\rm aux}(\textbf{x}) | \textbf{x} \in D_{\rm target}\}$
\STATE Derive a similarity measure, $S^{y_1}$ for all $\textbf{x}$ with the same $y_1$ for each $y_1$
\STATE If $D_{\rm aux}$ is available, train a classifier $f$ that predicts $y_2$ based on $\{(\textbf{x},y_1,S^{y_1},y_2)| \textbf{x} \in D_{\rm aux}\}$. 
Else, fit $\{(\textbf{x},y_1,S^{y_1})| \textbf{x} \in D_{\rm target}\}$ with a distribution $f'$
\RETURN $\{\hat{y_2} = f(\textbf{x},y_1,S^{y_1})|\textbf{x} \in D_{\rm target}\}$ or the likelihood score of $f'$, $\{{\mathcal L}(f'|\textbf{x},y_1,S^{y_1})|\textbf{x} \in D_{\rm target}\}$
\end{algorithmic}
\end{algorithm}

\section{Membership inference: evaluation}
\label{sec:eva}
We evaluate empirically the membership inference techniques discussed above applied to the three face recognition student models.
For each of the models, we first describe the evaluation setups, followed by evaluation with the first-cut solution. 
Then, we perform evaluation with class-based inference, introducing new techniques when necessary.
At the end of the Section, we give an overall discussion of our results.

\subsection{DNN features}
\label{sec:feat}
\label{subsc:feat_agg}

We assume that the adversary has access only to the output values of the feature extractor. 
Let $X$ be the space of the image. Then, the feature extractor may be described as a function with the following mapping: $\psi: X \to \mathbb{R}^k$, assuming that the feature vector is of dimension $k$.

$D_{\rm target}$ is set up as follows. 
For all VGGFace2 training subset face images, $\textbf{x}$ of class $y_1$, i.e., 
\begin{eqnarray*}
    (\textbf{x}, y_1) \in \mathcal{D}_{{\rm VGG}_{\rm train}}, 
\end{eqnarray*}

we sample a subset of them and label them as $y_2 = 1$, denoting the data subset by $D_{\rm member}$. That is, for all sampled $y_1$:  
\begin{eqnarray}
\label{eq:y2_1}
(\textbf{x}, y_1, y_2 = 1) \in D_{\rm member} \subseteq \mathcal{D}_{{\rm VGG}_{\rm train}}.
\end{eqnarray}

We have sampled in total 200 classes, and 50 face images from each class.

We further sample a subset of VGGFace2 test dataset and label them as $y_2 = 0$, denoting the data subset by $D_{\rm non-member}$. That is, for all sampled $y_1$:  
\begin{eqnarray}
\label{eq:y2_2}
(\textbf{x}, y_1, y_2 = 0) \in D_{\rm non-member} \subseteq \mathcal{D}_{{\rm VGG}_{\rm test}}.
\end{eqnarray}
We have also sampled in total 200 classes, and 50 face images from each class for $D_{\rm non-member}$.

\paragraphb{First-cut solution.}
The first-cut strategy of leaking information is simply feeding the feature (output of $\psi$) to a binary classifier to classify whether the data point belongs to the training data of the teacher model or not.

The adversary is assumed to have obtained the auxiliary knowledge, $(\textbf{x}, y_1, y_2) \in D_{\rm aux}$.
Using $ D_{\rm aux}$, the adversary can train a supervised attack model to infer whether a data point from $D_{\rm target}$ belongs to the training dataset or not with the output of $\psi$ as the input feature of the attack model. 

Equations \ref{eq:y2_1} and \ref{eq:y2_2} are the datasets to be fed to our ML algorithms to differentiate between $D_{\rm non-member}$ and $D_{\rm member}$. We use three supervised ML algorithms (linear model, random forest, linear support vector machine) to build the attack model and perform a 5-fold cross validation (i.e., out of the 5 subsets of the original $D_{\rm target}$, four of them are $D_{\text{aux}}$, while the remaining one is to be tested for its membership). The average results over 5-fold cross validation are shown in Table \ref{tab:naive_feature}.

In addition to using the original feature of dimension 512, we perform principal component analysis (PCA) to reduce the dimension to 50 (which captures more than 99\% of the variance) before applying ML algorithms on the dataset.
The results do not differ much from those shown in Table \ref{tab:naive_feature} however.

\begin{figure*}[ht]
\begin{subfigure}{.33\textwidth}
  \centering
  \includegraphics[width=.8\linewidth]{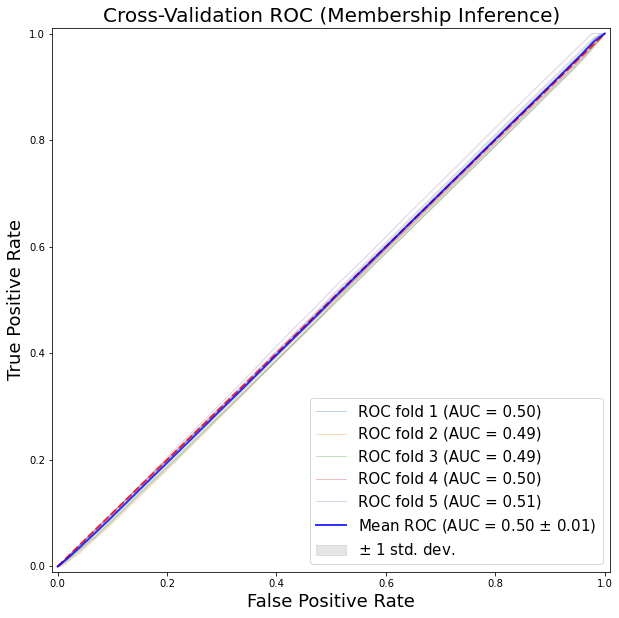}  
  \caption{DNN features}
  \label{fig:naive_feat_roc}
\end{subfigure}
\begin{subfigure}{.33\textwidth}
  \centering
  \includegraphics[width=.8\linewidth]{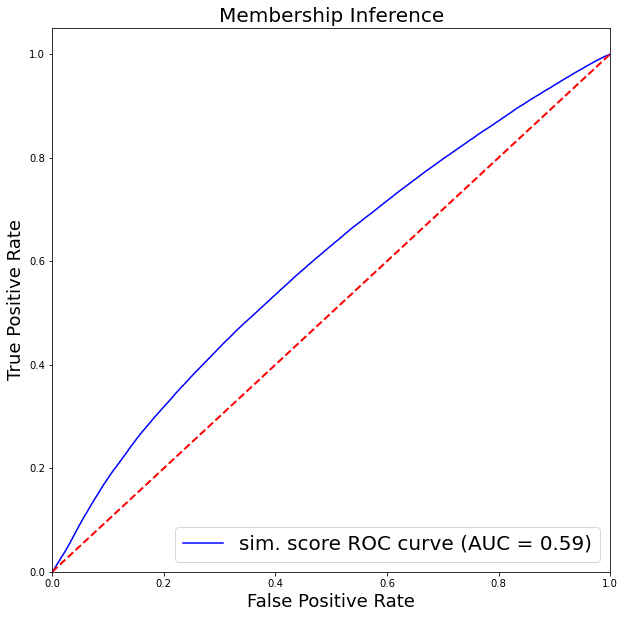}  
  \caption{Face verification.}
  \label{fig:naive_veri}
\end{subfigure}
\begin{subfigure}{.33\textwidth}
  \centering
  \includegraphics[width=.8\linewidth]{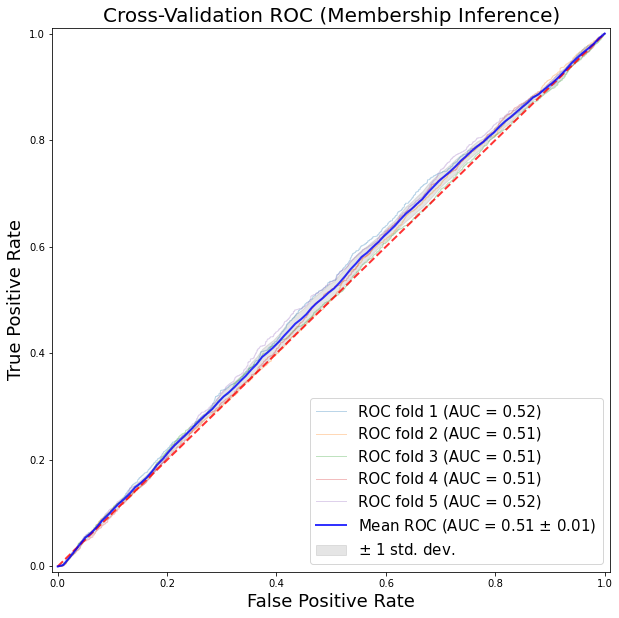}  
  \caption{Fine-tuned face verification}
  \label{fig:naive_face}
\end{subfigure}
\caption{First-cut membership inference results. In (a) and (b),the outputs obtained are used to train a classifier to perform the inference, assuming auxiliary information is available. In (b), the similarity score is used as the decision threshold to infer the membership without auxiliary information. The first-cut technique is unable to produce meaningful inferences with face verification as the only exception.}
\label{fig:naive}
\end{figure*}

As can be from the Table, the metrics are all less than 0.5. 
We also show the ROC curve in Figure \ref{fig:naive_feat_roc} by performing 5-fold cross validation and fitting the logistic regression model, which yields AUC $= 0.52$. 
This shows that the performance is not much better than random guessing.

\begin{table}[t]
\centering
\caption{Membership inference results on DNN features as a first-cut solution, using machine learning algorithms linear model (LM), random forest (RF), and linear support vector machine (SVM). Shown are average values and errors after performing 5-fold cross-validation.}
\label{tab:naive_feature}
\begin{tabular}{|l|c|c|c|}
\hline 
 ML & Precision &Recall & F1  \\ 
\hline 
LM   & $0.497 \pm 0.005$ &$0.467 \pm 0.069$& $0.479 \pm 0.035$\\
RF& $0.496 \pm 0.005$ &$ 0.470 \pm 0.007$& $0.493\pm 0.011$\\ 
SVM  & $0.497 \pm 0.005$ &$0.468 \pm 0.069$& $0.479 \pm 0.035$ \\
\hline
\end{tabular}
\end{table}

\paragraphb{Class-based inference.}
In order to perform class-based inference, the adversary first deduces the class or identity of the face images, i.e., $y_1$ in Equations \ref{eq:y2_1} and \ref{eq:y2_2}.
Then, the adversary infer the membership of instances in $D_{\rm target}$ of which the membership is unknown, albeit the class or identity can be correctly deduced (infer $y_2$ while $y_1$ is known).

An aggregate-level similarity measure $S^{y_1}$ for each identity/class $y_1$ is needed to perform class-based inference.
Here, we propose to construct $S^{y_1}$ based on the covariance matrix as it is a natural choice of similarity measure to quantify and capture the ``concentration" property of the instances.

Let $\{\phi_{y_1}\}^{m}$ be the $m$ feature vectors of class $y_1$ as a result of $\psi$ being fed with $m$ images of class $y_1$.
The covariance for each class can then be calculated based on the aggregated $m$ samples.

Denote the dimension of the output vector $\phi_{y_1}$ by $k$. Given face images from an identity, $y_1$, we construct a $k$-dimensional covariance matrix, $\sigma^{y_1}_{ij}$, where $i,j$ are the $i$-,$j$-th dimension of $\phi_{y_1}$.
Then, we take the $l_{\rho}$-norm sum over the weighted elements of the covariance matrix:
\begin{eqnarray}
    S^{y_1} &=& \sum_{i}^k|\sigma^{y_1}_{ii}|_{\rho} + \lambda \sum_{i\neq j}^k\frac{1}{2}|\sigma^{y_1}_{ij}|_{\rho},
\label{eq:cov}
\end{eqnarray}
where $\lambda$ is the weight parameter. $\rho = 2$ indicates the Euclidean sum, and $\lambda$ controls the weight between the diagonal and non-diagonal elements of $\sigma^{y_1}$.

\begin{figure*}[ht]
\begin{subfigure}{.33\textwidth}
  \centering
  \includegraphics[width=.8\linewidth]{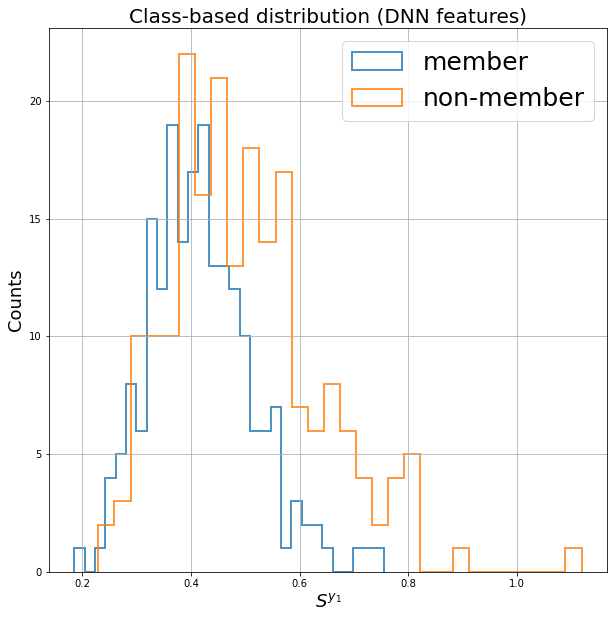}  
  \caption{DNN features}
  \label{fig:featdist}
\end{subfigure}
\begin{subfigure}{.33\textwidth}
  \centering
  \includegraphics[width=.8\linewidth]{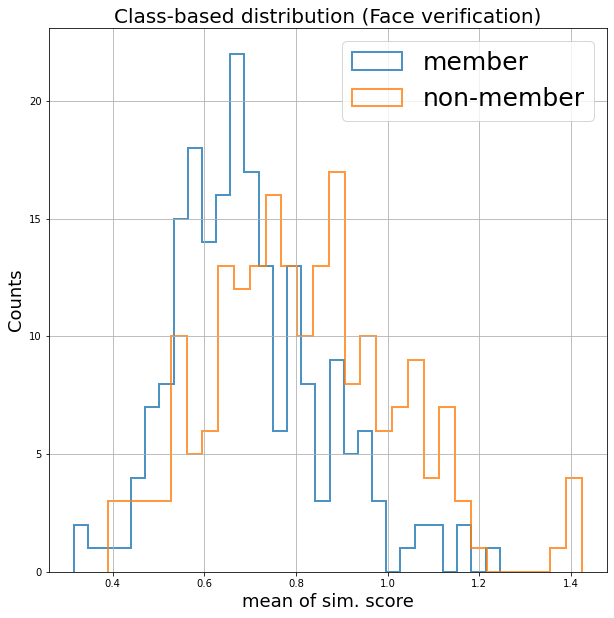}  
  \caption{Face verification}
  \label{fig:veridist}
\end{subfigure}
\begin{subfigure}{.33\textwidth}
  \centering
  \includegraphics[width=.8\linewidth]{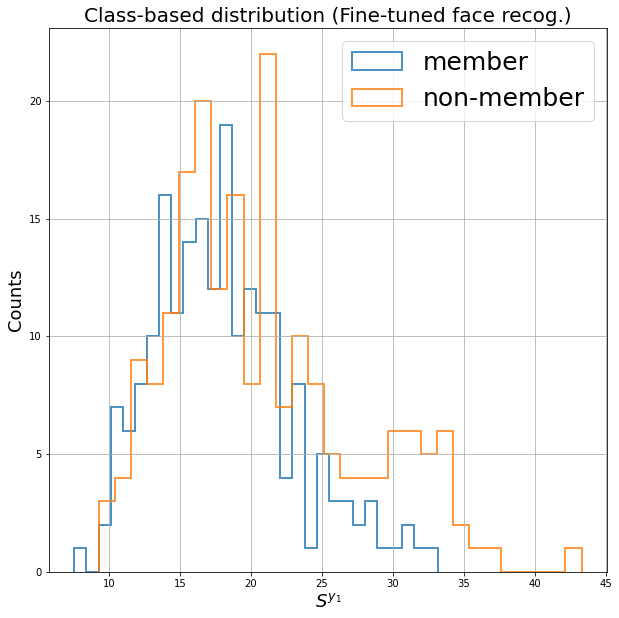}  
  \caption{Fine-tuned face recognition}
  \label{fig:facedist}
\end{subfigure}
\caption{Distribution of class-based variables for classes members ($y_2$=1, blue) and non-members ($y_2$=0, orange) of the teacher training dataset. This difference in distribution is exploited to distinguish members from non-members of the teacher model.}
\end{figure*}

In Figure \ref{fig:featdist}, we show the distribution of $S^{y_1}$ for member ($y_2$=1) and non-member ($y_2$=0) of the teacher training dataset, for $\rho=\lambda=1$.
As can be seen from the Figure, $D_{\rm member}$ has larger mean and spread in the covariance distribution.
This difference is to be exploited to perform inference.

\begin{figure*}[ht]
\begin{subfigure}{.33\textwidth}
  \centering
  \includegraphics[width=.8\linewidth]{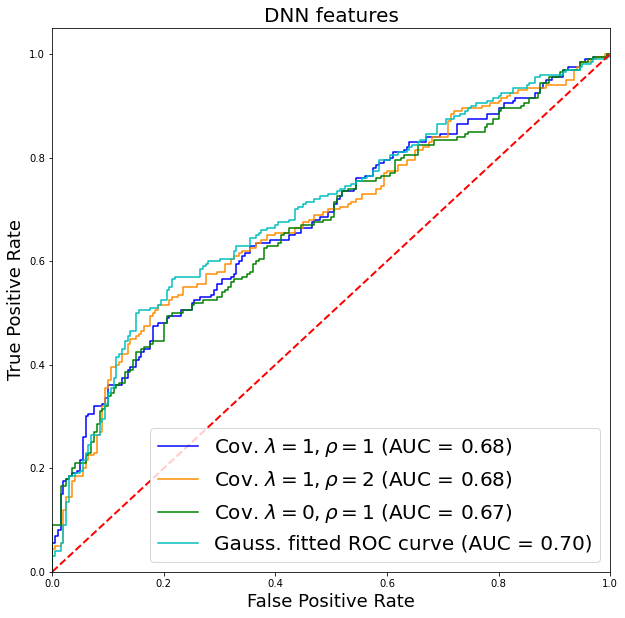}  
  \caption{DNN features}
  \label{fig:featroc}
\end{subfigure}
\begin{subfigure}{.33\textwidth}
  \centering
  \includegraphics[width=.8\linewidth]{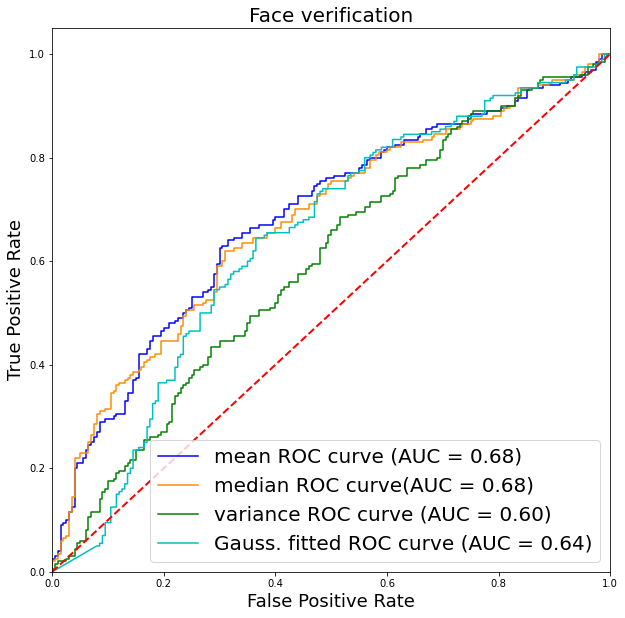}  
  \caption{Face verification}
  \label{fig:veriauc}
\end{subfigure}
\begin{subfigure}{.33\textwidth}
  \centering
  \includegraphics[width=.8\linewidth]{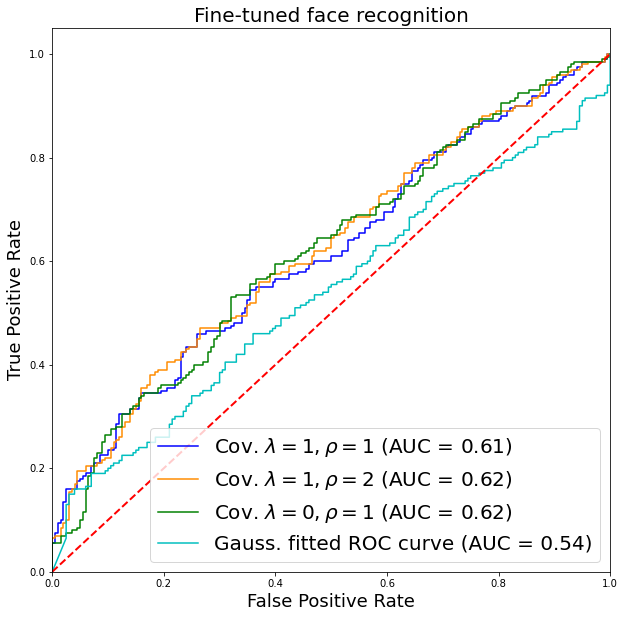}  
  \caption{Fine-tuned face recognition}
  \label{fig:faceroc}
\end{subfigure}
\caption{Attack results using our proposed class-based inference, using various measures, assuming that the adversary has no auxiliary knowledge. Improvement of inference results can be seen in all cases compared to, e.g. Figure \ref{fig:naive} where first-cut approaches are utilized.}
\end{figure*}

Let us assume that the adversary does not have information about $D_{\rm aux}$, and wishes to sort the instances of the target dataset by the membership prediction confidence. 
The discriminative power of $S^{y_1}$ is visualized by plotting the ROC curve using $S^{y_1}$ as the discrimination threshold.
The results varying $\rho$ and $\lambda$ are shown in Figure \ref{fig:featroc}, and it is noted that an AUC of 0.68 at best is achieved, an improvement compared to the first-cut membership inference technique.

Motivated by the fact that the member/non-member features are approximately normally distributed, as can be observed from Figure \ref{fig:featdist}, we also use the Gaussian mixture model (with 2 mixture components) to fit the distribution.
Before doing so, we need to define the input features of the Gaussian mixture model.

We first compute $\sigma^{y_1}$ as before. The upper triangular elements of $\sigma^{y_1}$ (including the diagonal elements) are extracted and treated as the features of a class ($k^2/2 + k/2$ in total).
Then, we apply PCA to reduce the feature dimension to 50.

After fitting with the Gaussian mixture distribution, we use the estimated posterior probability as the decision threshold and plot the ROC curve.
As can be observed in Figure \ref{fig:featroc}, an AUC of 0.70 is achieved, further improving the inference performance.

 We also assess the adequacy and the discriminative power of our approach of performing member/non-member classification, assuming that $D_{\rm aux}$ is available and the adversary wishes to perform supervised learning.
 We choose to run a quadratic discriminant analysis, motivated by the Gaussian-like distributions.

 Discriminant analysis is usually used to gauge how discriminative the features are, and the quadratic discriminant analysis fits a Gaussian density to each class without assuming that all classes share the same covariance matrix (in contrast to linear discriminant analysis, which assumes that all classes share the same covariance matrix).
 We use the same features utilized to fit the Gaussian mixture model mentioned above to perform the analysis.

\begin{figure*}[ht]
\begin{subfigure}{.33\textwidth}
  \centering
  \includegraphics[width=.8\linewidth]{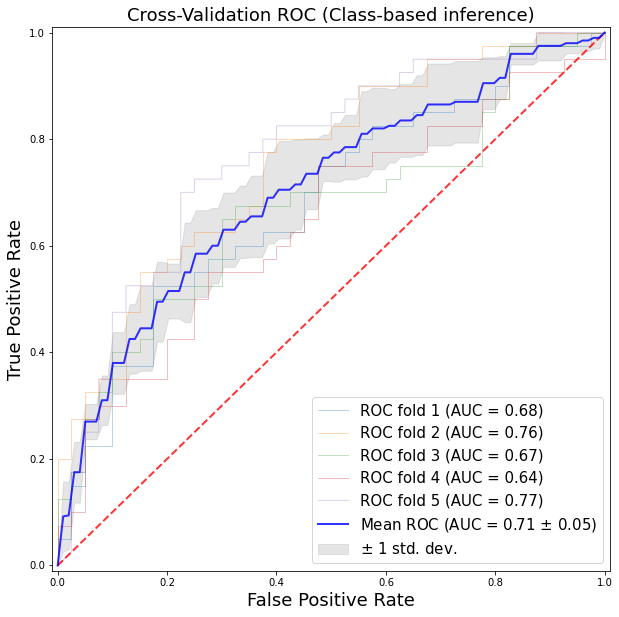}  
  \caption{DNN features}
  \label{fig:featcv}
\end{subfigure}
\begin{subfigure}{.33\textwidth}
  \centering
  \includegraphics[width=.8\linewidth]{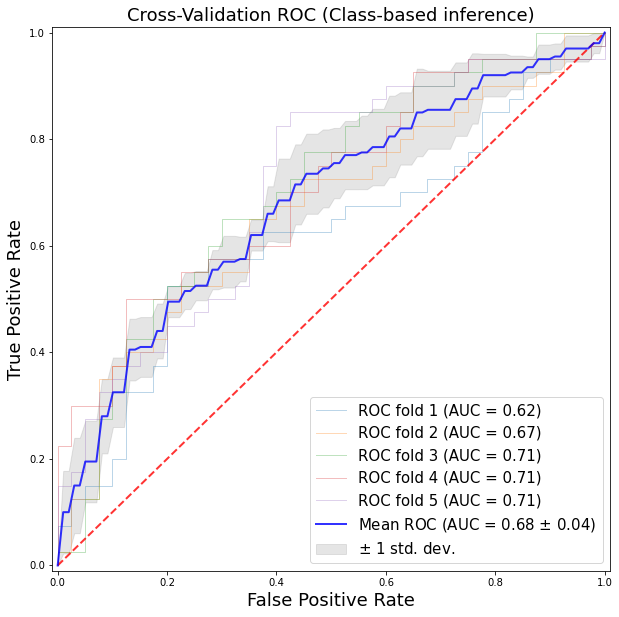}  
  \caption{Face verification}
  \label{fig:vericv}
\end{subfigure}
\begin{subfigure}{.33\textwidth}
  \centering
  \includegraphics[width=.8\linewidth]{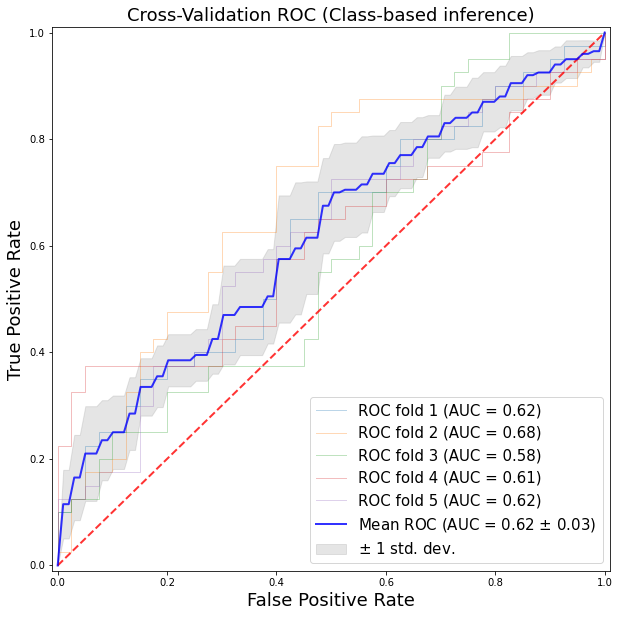}  
  \caption{Fine-tuned face recognition}
  \label{fig:facecv}
\end{subfigure}
\caption{Attack results using our proposed class-based inference, assuming that the adversary has auxiliary knowledge. The adversary trains a classifier with the auxiliary information to launch the attack. Results are presented in a 5-fold cross-validated way. The inference results are better than those presented in Figure \ref{fig:naive} utilizing first-cut approaches for all cases.}
\end{figure*}

The performance of the quadratic discriminant analysis is gauged with 5-fold cross validation (again, this means that, out of the 5 subsets of the original $D_{\rm target}$, four of them are $D_{\text{aux}}$, while the remaining one is to be tested for its membership) . As can be observed in Figure \ref{fig:featcv}, an average AUC of 0.71 is obtained. This is also an improvement compared to the first-cut solution.

\subsection{Face verification}
\label{sec:veri}
We next move to analyzing the scenario where the adversary has access to the output of a face verification system.

Let $X$ be the space of the image. Then, the face verification system may be described by a function with the following mapping: $\psi: X \times X \to \mathbb{R}$.

The adversary is able to query a pair of face images of her choice and observes the output.
We assume that the adversary makes the following queries: all combinations of image pair of the same identity are queried and their outputs are observed and recorded by the adversary. 

The dataset about which the properties are to be inferred are as follows. The subset belonging to the training dataset used to train the teacher model is, for all pairs $\textbf{x}_1,\textbf{x}_2$ sampled class/identity $y_1$, represented as:
\begin{eqnarray}
(\textbf{x}_1,\textbf{x}_2, y_1, y_2 = 1) 
\in D_{\rm member} 
\subseteq \mathcal{D}_{{\rm VGG}_{\rm train}},
\end{eqnarray}
denoted by $D_{\rm member}$.
The subset not belonging to the training dataset used to trained the teacher model is denoted by $D_{\rm non-member}$: 
\begin{eqnarray}
(\textbf{x}_1,\textbf{x}_2, y_1, y_2 = 0) \in D_{\rm non-member} \subseteq \mathcal{D}_{{\rm VGG}_{\rm test}}.
\end{eqnarray}
We have sampled 200 classes (with equal ratio of gender), and 50 face images for each class, for each of $D_{\rm member}$ and $D_{\rm non-member}$.
All possible pairs of $\textbf{x}_1,\textbf{x}_2$ from the 50 face images for each class ($50 \times 49/2 = 1225$ pairs) are considered.
The score or distance obtained by the adversary between $\textbf{x}_1$ and $\textbf{x}_2$ whose labels, $y_1$ are equal, are denoted $\tau^{y_1}_{\textbf{x}_1,\textbf{x}_2}$.

\paragraphb{First-cut solution.}
As a first approach of inferring the membership (assuming that the adversary does not have auxiliary knowledge), we use the output, $\tau^{y_1}_{\textbf{x}_1,\textbf{x}_2}$ as the discrimination threshold to differentiate between $D_{\rm non-member}$ and $D_{\rm member}$.  
The ROC curve is shown in Figure \ref{fig:naive_veri}, where an AUC of 0.59 is achieved.

It is noted that nothing much else can be inferred (e.g., building a classifier to perform inference) using only the individual instance even if the adversary has auxiliary knowledge, which is a collection of data with one-dimensional feature.

\paragraphb{Class-based inference}.
We seek a similarity function, $g$, which maps all members in a class to a value such that the membership can be better inferred: $g: \mathbb{R}^m \to \mathbb{R}$, given $m$ members in a class.

The most natural choice of $g$ is mean, i.e., we calculate $\tau^{y_1}$, by taking the average over all $\tau^{y_1}_{\textbf{x}_1,\textbf{x}_2}$ for all $\textbf{x}_1,\textbf{x}_2$ pairs belonging to $y_1$.
For each of $D_{\rm member}$ and $D_{\rm non-member}$, we plot the distribution of the mean of $\tau^{y_1}$ in Figure \ref{fig:veridist}. 
Again, $D_{\rm member}$ has smaller mean and spread in distribution, as the instances of $D_{\rm member}$ within the same class are trained to have small $\tau^{y_1}$. 

We also plot the ROC curves, using the mean, median and variance of  $\tau^{y_1}$ as the decision thresholds, assuming that the adversary does not have auxiliary knowledge.
As can be seen in Figure \ref{fig:veriauc}, an AUC of 0.68 can be achieved, improving the first-cut membership inference result.

Moreover, we have attempted to fit the distribution with the Gaussian mixture model.
We find that fitting the distribution with 3 mixture components, and using a combination of mean, median, variance, mean absolute deviation, median absolute deviation, and inter-quartile range as the input features give the optimal AUC (0.64).
Though, as can be seen in Figure \ref{fig:veriauc}, it seems that simply using the mean as the decision threshold gives the best inference result. 

Assuming that the adversary has auxiliary information, a classifier can be trained based on the aggregate-level information to perform inference attack.
The features used are the same as those used to fit the Gaussian mixture model.
Instead of the quadratic discriminant analysis, we find that logistic regression gives a better result.
The ROC plot with 5-fold cross validation is shown in Figure \ref{fig:vericv}, where the AUC is 0.68 for logistic regression (AUC of 0.65 is obtained using the quadratic discriminant analysis).

\subsection{Fine-tuned face recognition}
\label{sec:face}
\label{subsec:face_class}
Let $X$ be the space of the image. Then, the fine-tuned face recognition system may be described as a function with the following mapping: $\psi: X \to \mathbb{R}^c$, where $c$ is the number of identity/class.
The adversary simply queries the face images of $D_{\rm target}$ to obtain $c$ confidence scores corresponding to the identities/classes used to train the student model for each query.

The same setup of $D_{\rm target}$ described in Section \ref{sec:feat} is used.
We note that, the identities/classes used to train the student model and $D_{\rm target}$ are disjoint.

\paragraphb{First-cut solution.}
Assuming that the adversary has auxiliary knowledge, the straightforward way of performing membership inference is treating the $c$ class confidence scores (output of $\psi$) as the features to be fed to a binary classifier to classify whether the data point belongs to the training data of the teacher model or not.

The 5-fold cross validation ROC curve is also shown in Figure \ref{fig:naive_face}. 
The AUC is 0.51, which means that the classifier is not much better than random guessing.

For the case when auxiliary information is available, we run a 5-fold cross validation to test the inference performance. As shown in Table \ref{tab:naive_face}, the metrics are overall less than $\sim 0.5$.

\begin{table}[t]
\centering
\caption{Membership inference results on fine-tuned face recognition as a first-cut solution, using machine learning algorithms linear model (LM), random forest (RF), and linear support vector machine (SVM). Shown are average values and errors after performing 5-fold cross-validation.}
\label{tab:naive_face}
\begin{tabular}{|l|c|c|c|}
\hline 
 ML & Precision &Recall & F1  \\ 
\hline 
LM  & $0.498 \pm 0.006$ &$0.444\pm 0.087$& $0.465 \pm 0.045$\\
RF& $0.496 \pm 0.005$ & $0.499 \pm 0.014$ & $0.497 \pm 0.007$\\ 
SVM  & $0.498\pm 0.007$ &$0.444 \pm 0.088$& $0.465\pm 0.046$\\
\hline
\end{tabular}
\end{table}
\paragraphb{Class-based inference}.
We seek a function that maps the confidence scores within a class to a similarity measure, i.e., $g: \mathbb{R}^{c\times m} \to \mathbb{R}$, given $m$ members in a class.

We learn from Section \ref{subsc:feat_agg} that the measure based on the covariance matrix, $S^{y_1}$ is helpful for inference.
Here, instead of constructing  $S^{y_1}$ using the feature vectors as done above Equation \ref{eq:cov}, we construct the covariance matrix using the $c$ classes as the matrix elements.
The distribution of $S^{y_1}$ with $\lambda=\rho=1$ is shown in Figure \ref{fig:facedist}.

To demonstrate the effectiveness of using $S^{y_1}$ assuming that the adversary does not have auxiliary information, we plot the ROC curve with various values of $\rho$ and $\lambda$ as defined in Equation \ref{eq:cov}.
The results are shown in Figure \ref{fig:faceroc}.

In the same Figure, we also show the result of applying PCA to the data to reduce the dimension to 15 and fitting the Gaussian mixture model (with 3 mixture components) following the procedures described in Section  \ref{subsc:feat_agg}. 
Using $S^{y_1}$ yields better results however.

Finally, assuming that auxiliary information is available, a classifier is trained based on the aggregate-level information to perform inference attack.
The quadratic discriminant analysis is applied, and features used are the same as those used to fit the Gaussian mixture model.
As can be observed from Figure \ref{fig:facecv}, an average AUC of 0.62 is achievable over 5-fold cross-validation.

\subsection{Discussion}
Overall, we have shown that class-based inference attacks are effective at inferring training data of the teacher model, even when the adversary interacts solely with the student models.
On the other hand, the first-cut solutions are largely ineffective. 
Even for the face verification model where the first-cut solution is able to infer meaningfully (AUC=0.59), our proposed approach is capable of further improving the inference result to AUC=0.68.

The API that exposes the DNN features are the most vulnerable one to our attacks, followed by face verification.
The fine-tuned face recognition API is least vulnerable. 
This is perhaps not surprising, as the DNN features expose the largest amount of information (512-dimensional vector per instance) to the adversary, while the fine-tuned face recognition API reveals only 20 confidence scores in our study.

\section{Attribute inference}
\label{sec:attr}
Teacher models are designed to output features rich in semantic information about the input data to the student models.
If the input data is user-related, the extracted feature may capture information related to users not intended to be exposed.

As a motivating example, consider a service where, in order to protect the user's privacy, the user is required to upload the extracted feature (derived from the teacher model loaded on the user's device), instead of the raw data, to the cloud-based host.
However, it is plausible to think of scenarios where the feature data are intercepted by the adversary, and exploited to infer the private attributes (gender, race) of the user.

As we assume that the adversary is able to access the auxiliary dataset, the adversary can use it to learn a supervised classifier to perform inference.
We focus on the case where the available auxiliary information is limited. 
This is to reflect the realistic scenario where collecting auxiliary data is difficult, and to demonstrate that privacy leakage is possible even with a little amount of information.

The general approach to performing attribute inference is presented in Algorithm \ref{alg:attr}.
In the following, we treat gender as the sensitive attribute, and assume that the adversary intends to infer it from the student models.

\begin{algorithm}[t]
\caption{Attribute inference}
\begin{algorithmic}[1]
\label{alg:attr}
\STATE \textbf{Input:} target dataset $\textbf{x} \in D_{\rm target}$ , target model $\psi$,  auxiliary data $D_{\rm aux}$ labeled with sensitive attribute $s$
\STATE Obtain pairs of API output and sensitive attribute, $\{(\psi(\textbf{x}),s) | \textbf{x} \in D_{\rm aux}\}$
\STATE Train a classifier $f$ that predicts $s$ based on $\{(\psi(\textbf{x}),s)\}$
\RETURN $\{\hat{s} = f(\psi(\textbf{x}))|\textbf{x}\in D_{\rm target}\}$
\end{algorithmic}
\end{algorithm}

\subsection{DNN features}
\label{subsec:feat_attr}

\begin{figure*}[ht]
\begin{subfigure}{.33\textwidth}
  \centering
  \includegraphics[width=.8\linewidth]{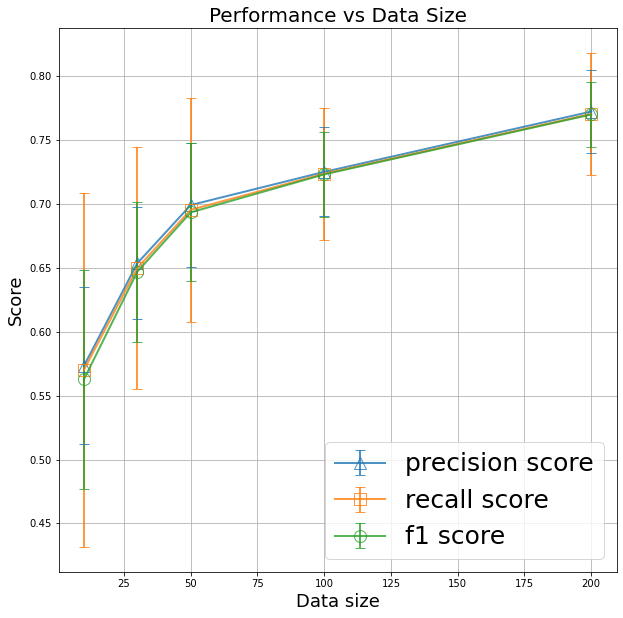}  
  \caption{DNN features}
  \label{fig:feat_attr_data}
\end{subfigure}
\begin{subfigure}{.33\textwidth}
  \centering
  \includegraphics[width=.8\linewidth]{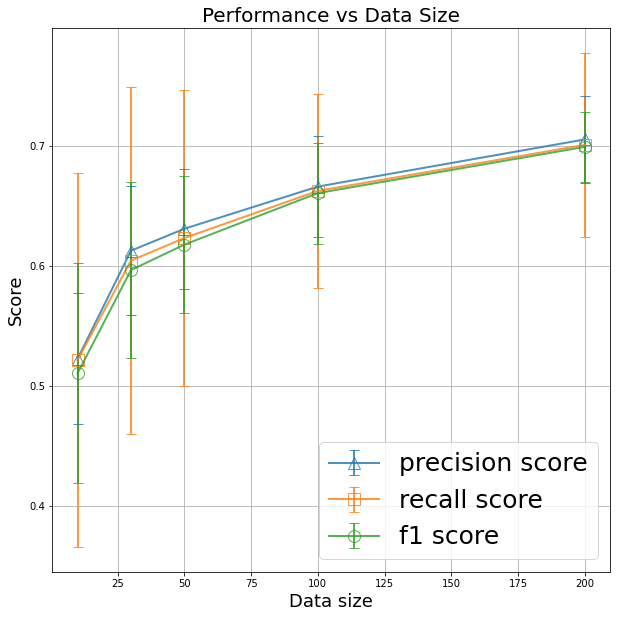}  
  \caption{Face verification}
  \label{fig:veri_attr_data}
\end{subfigure}
\begin{subfigure}{.33\textwidth}
  \centering
  \includegraphics[width=.8\linewidth]{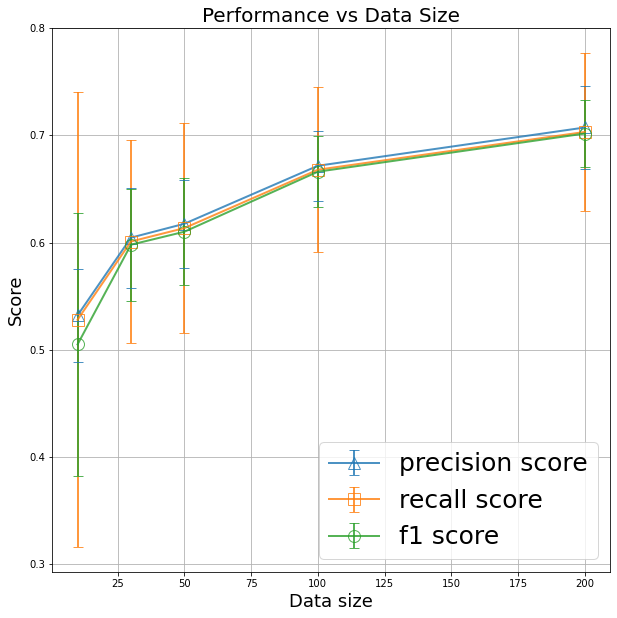}  
  \caption{Fine-tuned face recognition}
  \label{fig:face_attr_data}
\end{subfigure}
\caption{Attribute inference performances versus auxiliary data size. 
As expected, the inference performance drops with an decrease of auxiliary data but the result is still better than random guessing with as few as 30 (50) labeled instances for case a and c (case b). }
\label{fig:attr}
\end{figure*}

If the adversary is able to obtain the feature vector, she can perform attribute inference rather straightforwardly, i.e., by using transfer learning techniques.
We use a linear classifier for inference.

The settings are as follows. 
2000 identities (equal gender ratio) are sampled from $\mathcal{D}_{{\rm VGG}_{\rm test}}$ to serve as samples in $D_{\rm target}$ and $D_{\rm aux}$.

The number of data points in $D_{\rm target}$ is fixed to be 400. We vary the number of data points in $D_{\rm aux}$ to see how it affects the inference performance. We repeat our experiments in a 5-fold cross-validated way.
The result is shown in Figure \ref{fig:feat_attr_data}.

\subsection{Face verification}
As mentioned in Section \ref{sec:veri}, the face verification API interacts as follows: $\psi (\textbf{x}_1, \textbf{x}_2)= d \in \mathbb{R}$.
Without loss of generality, we assume that $\textbf{x}_2 \in D_{\rm aux}$ are the ``fixed" auxiliary samples. 

Let us further explain this setup.
The purpose of the adversary is to infer the attribute $s^*$ of $\textbf{x}^*\in D_{\rm target}$, given API response $\psi (\textbf{x}^*, \textbf{x}_2)$.
Auxiliary information is available as follows: $\{(\psi(\textbf{x}_1, \textbf{x}_2),s)\}$.

20 identities (equal gender ratio) are selected from $\mathcal{D}_{{\rm VGG}_{\rm test}}$ to serve as $ \textbf{x}_2$.
Additionally, 2000 identities (equal gender ratio) are sampled independently from $\mathcal{D}_{{\rm VGG}_{\rm test}}$ to serve as $ \textbf{x}_1$, i.e., samples of $D_{\rm target}$ and $D_{\rm aux}$.

The adversary queries each identity in $\textbf{x}_1$ to pair with all the identities in $\textbf{x}_2$.
The attribute inference problem may then be formulated as a problem of training a gender classifier $f: \mathbb{R}^{n_{\textbf{x}_2}} \to \mathbb{R}^2$, where $n_{\textbf{x}_2}$ is the number of identities in $ \textbf{x}_2$.

As in Section \ref{subsec:feat_attr}, we vary the number of data points in $D_{\rm aux}$, fixing the number of data points in $D_{\rm target}$ to be 400.
The result is shown in Figure \ref{fig:veri_attr_data}.

\begin{figure*}[ht]
\begin{subfigure}{.5\textwidth}
  \centering
  \includegraphics[width=.6\linewidth]{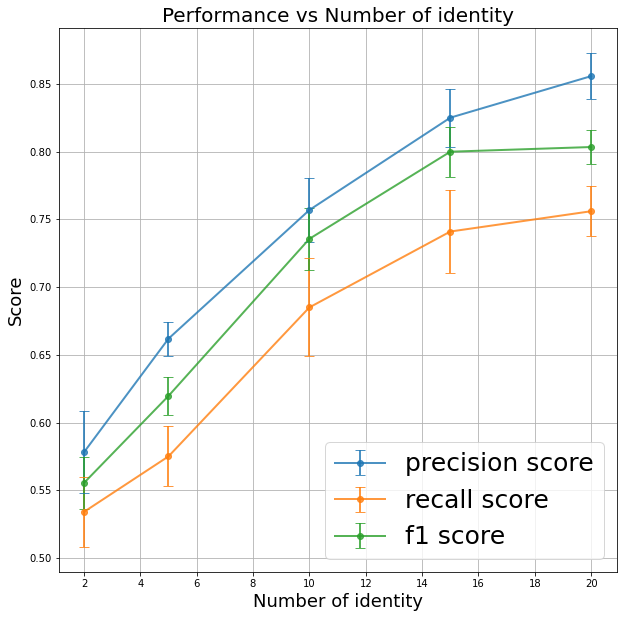}  
  \caption{Face verification}
  \label{fig:veri_attr_id}
\end{subfigure}
\begin{subfigure}{.5\textwidth}
  \centering
  \includegraphics[width=.6\linewidth]{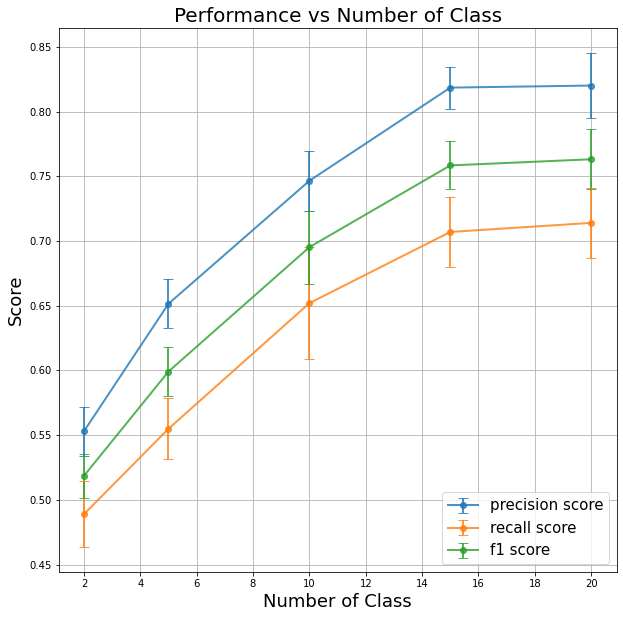}  
  \caption{Fine-tuned face recognition}
  \label{fig:face_attr_id}
\end{subfigure}
\caption{Attribute inference performance versus the amount of information exposed to the adversary. Inference is still meaningful with as few as 2 (5) identities (classes) for the face verification (fine-tuned face recognition) student model.}
\end{figure*}

In addition, we vary the number of $\textbf{x}_2$ to investigate how limited information may affect the performance, using the 5-fold cross-validation procedure mentioned above, showing the result in Figure \ref{fig:veri_attr_id}.

\subsection{Fine-tuned face recognition}
The fine-tuned face recognition API behaves as described in Section \ref{sec:face} as follows: $\psi (\textbf{x})= y \in \mathbb{R}^c$.
The adversary is assumed to know the identities/classes to be classified by the student model.

As in Section \ref{sec:face}, the student model is a 20-class face recognition model (the identities are sampled from $\mathcal{D}_{{\rm VGG}_{\rm test}}$).
Additionally, 2000 identities (equal gender ratio) are sampled independently from $\mathcal{D}_{{\rm VGG}_{\rm test}}$ to serve as samples in $D_{\rm target}$ and $D_{\rm aux}$.

The attribute inference problem may then be formulated as a problem of training a gender classifier $f: \mathbb{R}^{c} \to \mathbb{R}^2$, treating the class confidence scores as features.

As in Section \ref{subsec:feat_attr}, we vary the number of data points in $D_{\rm aux}$, fixing the number of data points in $D_{\rm target}$ to be 400.
The result is shown in Figure \ref{fig:face_attr_data}.

Furthermore, we vary the number of class to be classified by the student model, $c$, to investigate how limited information may affect the performance, using the 5-fold cross-validation procedure mentioned above.
Figure \ref{fig:face_attr_id} shows the result.

\paragraphb{Overall discussion.}
The performance drops appreciably when auxiliary data decrease in number, as shown in Figure \ref{fig:attr}.
Nevertheless, around 30 (50) labeled instances are enough to make meaningful inference on the DNN features and fine-tuned face recognition models (face verification model).
The performance also degrades when the query information per auxiliary data point becomes less.
Still, obtaining 2 similarity scores from the face verification API is enough to deduce the gender better than random guessing.
Correspondingly, a face recognition API with a minimum of 5 classes is sufficient to make meaningful inference.

\section{Defenses}
\label{sec:defense}
We focus on simple and low-cost mitigation strategies that do not require re-training or modification of the teacher model.
The reasons are the cost to re-train the teacher model is typically high, and re-deployment of the teacher model may not be desirable for models already put into production within the commercial setting.
Techniques that require re-training are left for future work.

\paragraphb{Output randomization.} The API responses are added with noises sampled from a normal distribution.
Let $x \in \mathbb{R}$ be the clean output. The resulting output after adding noise of scale $\eta$ distributes as $\mathcal{N}(x, \eta x)$. 
We add such noises to all elements of the output vector and normalize it appropriately.
Note that this reduces the API's utility.

\paragraphb{Rounding numbers.} The significant figures of the outputs are reduced to limit information exposed to the adversary. 

\paragraphb{Outputting top-$k$ predictions.} Only the predictions with the highest probabilities are revealed. This applies only to the fine-tuned face recognition student model.

\subsection{Results}
\paragraphb{Output randomization.} We experiment with various level of noise, $\eta$, to see how the performance class-based inference, measured with AUC, degrades (assuming that the adversary does not have auxiliary knowledge. See previous sections). The inference gets closer to purely random guessing as AUC approaches 0.5.
In Table \ref{tab:noise}, we show the results of adding noise to the three APIs in consideration.

\begin{table}[t]
\centering
\caption{Effects of output randomization (with scale $\eta$ as defined in text) on AUC added to DNN features (F.), face verification model (V.), and fine-tuned face recognition model (R.).}
\label{tab:noise}
\begin{tabular}{|l|c|c|c|}
\hline 
 $\eta$ & 0.1 & 1.0  & 5.0  \\ 
\hline 
F.  & $0.676\pm 0.001$& $0.674\pm0.003$ & $0.494\pm0.017$ \\
V.& $0.684 \pm 0.000$ & $0.684 \pm 0.003$& $0.645 \pm 0.013$\\ 
R.  & $0.621\pm0.002$ & $0.494\pm 0.006$  & $0.451 \pm 0.0139$\\
\hline
\end{tabular}
\end{table}

\paragraphb{Rounding numbers.}
We find that our inference attacks are robust against this strategy.
Even after rounding the numbers to 1 significant figure, all the AUCs tested remain the same (0.70, 0.68, 0.62 for DNN features, face verification, fine-tuned face recognition respectively).

\paragraphb{Outputting top-$k$ predictions.}
Since the API reveals only a limited number of class confidence score, the adversary is unable to construct the full $c$-dimensional covariance matrix $\sigma^{y_1}$ as described in Section \ref{subsec:face_class}.
We consider two scenarios: (1) constructing a $k$-dimensional covariance matrix when only top-$k$ classes are exposed; (2) constructing a $c$-dimensional covariance matrix, but replacing the unknown confidence scores with 0.

For (1), we find that, varying $k$ from 15 to 2, the AUC stays almost constant with an value of 0.57.
For the extreme case where $k=1$, the inference is close to random guessing (AUC=0.5034).
For (2),  varying $k$ from 15 to 1, the average AUC is $\sim 0.53$.

\paragraphb{Overall discussion.}
Inference attacks on the fine-tuned face recognition model can be mitigated with noise, albeit injecting noises means that confidence scores are lowered, i.e., it comes at a price of utility.
Limiting the number of class confidence score is partially successful, but inference is still possible. This is because information about \textit{wrong} predictions (which has larger variance if the data point is non-member) are captured by the covariance matrix and used to perform inference.
Inference attacks on DNN features and face verification API are quite robust in comparison. This is not surprising because they expose relatively larger amount of information to the adversary compared to the fine-tuned face recognition model.

\section{Related work}
\label{sec:related}
The study of privacy risks when releasing statistical information such as ML models can be traced to \cite{DBLP:conf/pods/DinurN03}, denoted \textit{reconstruction attack} by the authors.

\paragraphb{Membership Inference.}
Membership inference attacks on more complex ML models are shown to be possible more recently
\cite{DBLP:conf/sp/ShokriSSS17,DBLP:conf/csfw/YeomGFJ18,mlleaks,comprehensive,DBLP:conf/uss/LeinoF20}.
Membership and attribute inference attacks have also been studied under collaborative learning \cite{DBLP:conf/sp/MelisSCS19}, online learning \cite{updateleaks} and generative adversarial networks \cite{ganleaks} settings.

While completing this paper, we found two papers which also discuss membership inference within the transfer learning setting \cite{DBLP:journals/corr/abs-2009-01989,DBLP:journals/corr/abs-2009-04872}.
\cite{DBLP:journals/corr/abs-2009-01989} studied transfer learning paradigms different from ours, where the adversary has direct access to the teacher model.
\cite{DBLP:journals/corr/abs-2009-04872} reached the same conclusion as us, i.e., standard membership inference techniques applied to student model APIs are ineffective at inferring the membership of teacher models' training data. However, we additionally propose new solutions to overcome this.

\paragraphb{Privacy beyond membership inference.}
In addition to the membership inference, we have several adversarial attacks that predict private properties from ML models.
Model inversion is an adversarial attack that predicts private information used as the inputs of the model \cite{fredrikson2015model}, not only the presence or absence in the input which the membership inference assumes.
\cite{tramer2016stealing} introduced the
model stealing attack against black-box ML models.
In this attack, an adversary attempts to learn the target ML model’s parameters.
Similar to attribute inference, property inference is an attack that infers whether or not an model includes a particular property in the training inputs, such as the environment in which the data was produced  \cite{ganju2018property}.
ML models may also unintentionally memorize the whole training instance (e.g., sensitive text message)  \cite{DBLP:conf/uss/Carlini0EKS19}.

As for defense against those inference attacks, differential privacy \cite{dwork2006differential} provides provable guarantees against such adversarial behaviors.
DP-SGD \cite{abadi2016deep}, which crafts randomized gradients to update parameters of the model, is a well-known framework to make the model differentially private.
Furthermore, transforming the training inputs to the differentially private ones is another practical approach \cite{zhang2014privbayes,takagi2020p3gm}.

\paragraphb{Adversarial Examples.}
The phenomenon of adversarial examples in deep learning has attracted a lot of attention \cite{szegedy2014intriguing} \cite{goodfellow2014explaining}. 
Adversarial example is a potentially critical safety issues in machine learning based systems.
Adversarial examples within the image recognition domain are images added with imperceptibly small perturbations such that misclassification occurs to the ML classifier even though the perturbed image can be correctly classified by human. It was shown in \cite{DBLP:conf/ccs/SharifBBR16} that fooling face recognition systems deployed in the physical world is possible by adding perturbations in the eyeglass region.
\cite{DBLP:conf/pst/KakizakiY019} proposed adversarial examples against feature extractor of face recognition systems.
Studies about generating adversarial examples are important for evaluating the robustness of ML models \cite{goodfellow2014explaining,carlini2017towards,kurakin2016adversarial,eykholt2017robust,yakura2019robust,DBLP:conf/bigdataconf/Takahashi19,DBLP:conf/uss/DemontisMPJBONR19,DBLP:conf/uss/Tong0HXZV19,DBLP:conf/uss/SuyaC0020} .
Adversarial examples have also found usage in privacy protection, i.e., preventing one's face from being recognized \cite{DBLP:conf/uss/ShanWZLZZ20}.
\cite{jia2019memguard} introduced a defense method against black-box membership inference via adversarial examples.

Meanwhile, defense against adversarial examples is an active research area too.
One simple defense approach against adversarial examples is to mask gradients \cite{papernot2016distillation}, but has been shown to be ineffective \cite{carlini2017towards,athalye2018obfuscated}.
\cite{madry2017towards} proposed a robust training that injects adversarial examples with correct labels into training samples, also known as adversarial training, which is thus far one of the most robust defensive strategies.

\section{Conclusion}
\label{sec:conclusion}
In this paper, we have demonstrated, using face recognition as a concrete example, that transfer learning models are vulnerable to privacy attacks. 
Due to our innovation, i.e., the class-based inference, we are able to show that privacy leakage can occur even when the (teacher) model is not interacting with the adversary directly.

Dataset membership in the context of face recognition is highly personal and sensitive. 
Hence, our results have important impacts on privacy in practice.

Let us point out some future directions worth pursuing.
We have thus far only investigated fairly simple defensive strategies.
It is worthwhile to consider privacy-preserving methods with theoretical guarantees, e.g., differential privacy in the future.
Furthermore, only transfer learning in the context of face recognition is investigated in this work.
Privacy issues of transfer learning in other domains are worth studying too, e.g., neural language processing and reinforcement learning.

{\footnotesize \bibliographystyle{acm}
\bibliography{main}}


\end{document}